\def\year{2018}\relax
\documentclass[letterpaper]{article} 
\usepackage{aaai18}  
\usepackage{times}  
\usepackage{helvet}  
\usepackage{courier}  
\usepackage{url}  
\usepackage{graphicx}  
\usepackage{amsmath}
\usepackage{subcaption} 
\usepackage[ruled]{algorithm}
\usepackage{algpseudocode}
\usepackage{color}
\usepackage{tabularx}
\usepackage{soul}
\makeatletter
\newcommand*{\rom}[1]{\expandafter\@slowromancap\romannumeral #1@}
\newcommand{\name}[1]{MLDG}

\newcommand{\keypoint}[1]{\vspace{0.1cm}\noindent\textbf{#1}\quad}

\makeatother
 
\begin{document}
%
\title{Learning to Generalize: Meta-Learning for Domain Generalization}
\author{Da Li \quad Yongxin Yang \quad Yi-Zhe Song \quad Timothy M. Hospedales\\
Queen Mary University of London \quad
University of Edinburgh \\
{\tt\small \{da.li, yongxin.yang, yizhe.song\}@qmul.ac.uk, 
t.hospedales@ed.ac.uk}}
\maketitle
\begin{abstract}
Domain shift refers to the well known problem that a model trained in one source domain performs poorly when applied to a target domain with different statistics. {Domain Generalization} (DG) techniques attempt to alleviate this issue by producing models which by design generalize well to novel testing domains. We propose a novel {meta-learning} method for domain generalization. Rather than designing a specific model that is robust to domain shift as in most previous DG work, we propose a model agnostic training procedure for DG. Our algorithm  simulates train/test domain shift during training by synthesizing virtual testing domains within each mini-batch. The meta-optimization objective requires that steps to improve training domain performance should also improve testing domain performance.
This meta-learning procedure trains models with good generalization ability to novel domains. We evaluate our method and achieve state of the art results on a recent cross-domain image classification benchmark, as well demonstrating its potential on two classic reinforcement learning tasks. 

\end{abstract}

\section{Introduction}

Humans are adept at solving tasks under many different conditions. This is partly due to fast adaptation, but also to a lifetime of encountering new task conditions providing the opportunity to develop of strategies that are  robust to different task contexts. If a human discovers that their existing strategy fails in a new context they do not just adapt, but further try to update their strategy to be more context independent, so that next time they arrive in a new context they are more likely to succeed immediately. We would like artificial learning agents to solve many tasks under different conditions (domains) and similarly solve the second order task of constructing models that are robust to change of domain and perform well `out of the box' in new domains. For example we might like computer vision systems to recognise objects immediately and without retraining, when the camera type is changed \cite{patel2015vdaSurvey}, or reinforcement learning trained agents to perform well immediately when placed in a new environment or subjected to changed morphology \cite{taylor2009TL_RL_Survey} -- without waiting for adaptation.

Standard learning approaches tend to break down when applied in different conditions (ie to data with different statistics) than used for training. This is known as domain or covariate shift \cite{storkey2007covariateShift}, and seriously affects the usefulness of machine learning models as we do not always have access to training data that is exactly representative of the intended testing scenario. Approaches to addressing this issue can be categorized into \textit{domain adaptation} (DA) and \textit{domain generalization} (DG). DA is relatively well studied, and addresses using unlabeled or sparsely labeled data in the target domain to quickly adapt a model trained in a different source domain \textcolor{black}{\cite{patel2015vdaSurvey,csurka2017domainAdaptationBook}}. The less well studied DG addresses building models that by design function well even in new target/testing domains. In contrast to DA, a DG model is not updated after training, and the issue  is how well it works out of the box in a new domain. The few existing DG methods typically train on multiple source domains and propose mechanisms to extract some domain agnostic representation or model that describes common aspects of known domains \cite{khosla2012undoing,muandet2013domain,ghifary2015domain,da2017dg}. They assume that such a common factor among existing source domains will persist to new testing domains, and thus provide a basis for generalization. DG is a harder problem  than DA in that it makes fewer assumptions (target data not required) but for the same reasons, it may be more valuable if solved.


We take a  \textit{meta learning} approach to DG. Rather than proposing a specific model suited for DG \cite{khosla2012undoing,ghifary2015domain,da2017dg}, we propose a model-agnostic training algorithm that trains any given model to be more robust to domain shift. This is related to the long standing idea of \emph{learning to learn} \cite{thrun1998learntolearn,schmidhuber1997inductiveBias}, which has recently seen a resurgence of popularity with applications to few-shot learning \cite{finn2017model,ravi2016optimization} and learning optimizers \cite{andrychowicz2016learning}. The most related of these studies to ours is the MAML approach of \cite{finn2017model}. MAML takes a meta-learning approach to few-shot learning by training a single model on a set of source tasks that is only a few gradient descent steps away from a good task-specific model. This meta-optimization objective trains models suited for few-shot fine-tuning to new target tasks. The DG problem is different because we to transfer across domains rather than tasks, and because DG assumes \emph{zero}, rather than few training examples of the target problem. 

Our meta-learning domain generalization approach (\name) provides a model agnostic training procedure that improves the domain generality of a base learner. Specifically, \name~ trains a base learner on a set of source domains by synthesising virtual training and virtual testing domains within each minibatch. The meta-optimization objective is then: to minimise the loss on the training domains, while also ensuring that the direction taken to 
achieve this also leads to an improvement in the (virtual) testing loss. 
We present analyses that give various perspectives on this strategy, including as following an optimization trajectory where the virtual training and virtual testing gradients are aligned. Overall our \name~ approach has several key benefits: As a meta-learning procedure, it does not introduce any new parameters, unlike other model-based DG approaches that grow parameters linearly in the number of source domains \cite{khosla2012undoing,ghifary2015domain,da2017dg,bousmalis2016domain} resulting in large numbers of total parameters. Similarly \name~ does not place any constraint on the architecture of the base learner and moreover can be applied to both supervised and reinforcement learning; where \textcolor{black}{prior DG alternatives \cite{khosla2012undoing,ghifary2015domain,da2017dg} both constrain the model architecture and address supervised learning}.

To summarise our contributions: We develop a gradient-based meta-learning algorithm that trains models for improved domain generalisation ability. Our algorithm can train any type of base network and applies to both supervised and reinforcement learning settings. We evaluate our approach on a very recent cross domain image recognition benchmark and achieve state of the art results, as well as demonstrating its promising applicability to two classic reinforcement learning tasks.

\section{Related Work}
\vspace{-0.1cm}
\keypoint{Multi-Domain Learning (MDL)} MDL addresses training a \emph{single} model that is effective for multiple known domains \cite{daume2007easyDA,rebuff2017mdl,bilen2017universal}. Domain generalization often starts with MDL on some source domains  but addresses training a model that generalizes well to held out unknown domains.

\keypoint{Domain Generalization} Despite the variety of the different methodological tools, most existing DG methods are built on three main strategies. The simplest approach is to train a model for each source domain. When a testing domain comes, estimate the most relevant source domain and use that classifier \cite{xu2014lowRankLatentDomain}. A second approach is to presume that any domain is composed of an underlying globally shared factor and a domain specific component. By factoring out the domain specific and domain-agnostic component during training on source domains, the domain-agnostic component can be extracted and transferred as a model that is likely to work on a new source domain \cite{khosla2012undoing,da2017dg}. Finally, there is learning a domain-invariant feature representation. If a feature representation can be learned that minimises the gap between multiple source domains, it should provide a domain independent representation that performs well on a new target domain. This has been achieved with multi-view autoencoders 
\cite{ghifary2015domain} and mean map embedding-based techniques \cite{muandet2013domain}. It has also been achieved based on gradient reversal domain confusion losses in deep networks \cite{ganin2015unsupervised,bousmalis2016domain}. Here multiple source domains are trained with an additional multi-task loss that prefers a shared representation for which domains are indistinguishable. Although initially proposed for DA rather than DG, these approaches can be adapted to the DG setting \cite{da2017dg}. In contrast to these studies, ours is the first to addresses domain generalization via meta-learning.

\keypoint{Neural Network Meta-Learning} Meta-learning methods for neural networks have a long history \cite{thrun1998learntolearn,schmidhuber1997inductiveBias}, but have resurged in popularity recently. 
Recent meta-learning studies have focused on learning good weight initializations for few-shot learning \cite{finn2017model,parisotto2016transferRL}, meta-models that generate the parameters of other models \cite{vinyals2016oneShot,da2017dg}, or learning transferable optimizers \cite{ravi2016optimization,andrychowicz2016learning}. Our approach is most related to those that learn transferable weight initializations, notably MAML \cite{finn2017model}. In MAML a single shared source model shared is trained using multiple source tasks. The meta-learning process simulates transfer learning by fine-tuning, so the global model is updated to solve each source task in turn based on a few examples and a few gradient descent steps. By training the source model such that all simulated testing tasks fine-tune well, meta-learning produces a source model that is easy to adapt. Both MAML and our \name ~ are model agnostic in that they apply to any base architecture and both supervised and to reinforcement learning settings. However, MAML addresses few-shot transfer to new tasks, rather than zero-shot transfer to new domains. 
In our case a different meta-learning objective is necessary because in DG we will not have access to target examples for fine-tuning during the actual testing. Therefore we propose a new meta-learning objective based around simulating domain shift and training such that steps to improve the source domain also improve the simulated testing domains.


\section{Methodology}

\subsection{Meta-Learning Domain Generalization}

\begin{figure}[t]
\centering
\includegraphics[width=1.0\linewidth]{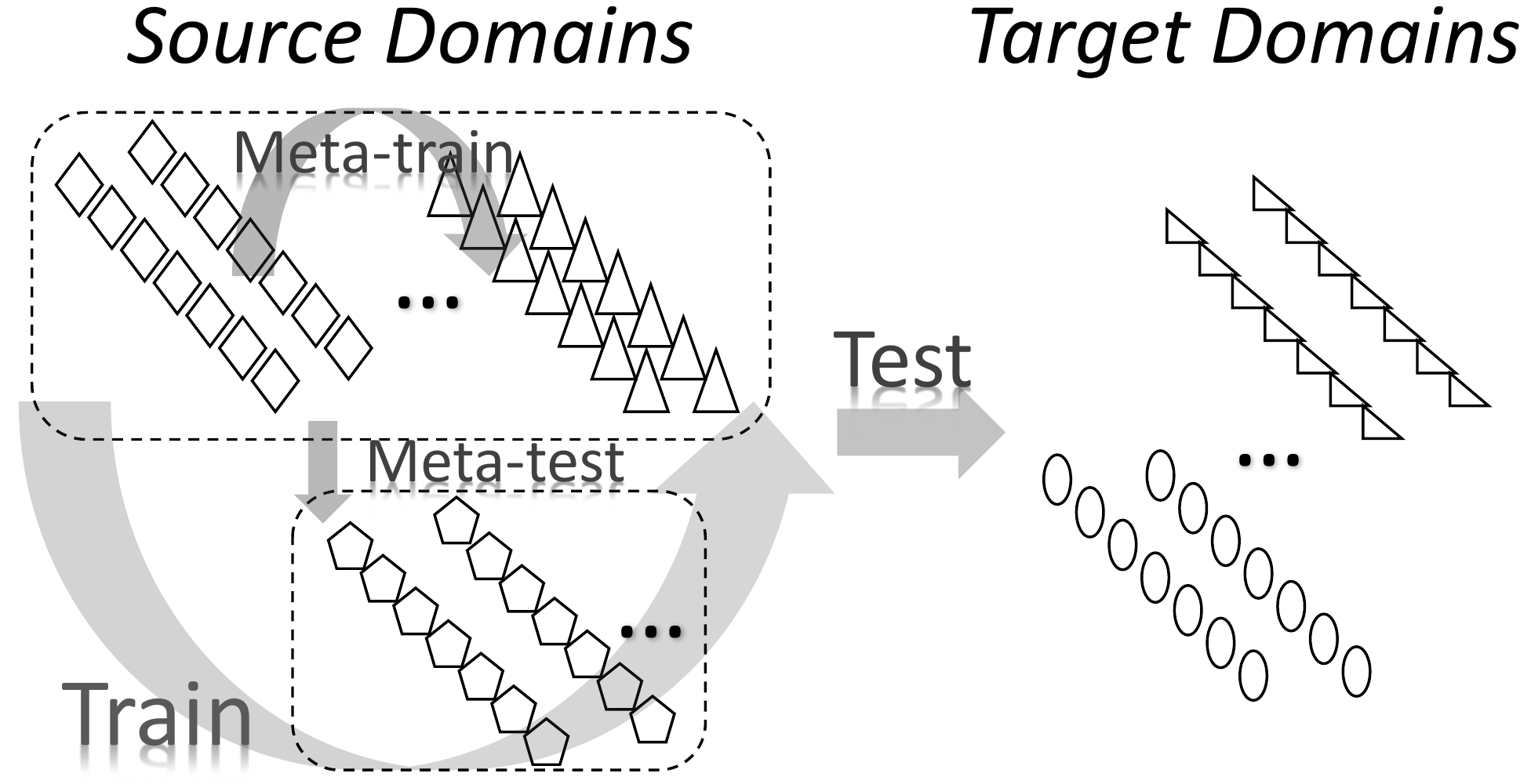}
\caption{Illustration of our Meta-Learning Domain Generalization method. Symbols represent different data domains.}
\label{illustration}
\end{figure}

In the DG setting, we assume there are $S$ source domains $\mathcal{S}$ and $T$ target domains $\mathcal{T}$. All of them contain the same task (same label space, and input feature space) but have different statistics. 
We define a single model parametrized as $\Theta$ to solve the specified task. DG aims for training $\Theta$ on the source domains, such that it generalizes to the target domains. To achieve this, at each learning iteration we split the original $S$ source domains $\mathcal{S}$ into $S-V$ \textbf{meta-train} domains $\bar{\mathcal{S}}$ and $V$ \textbf{meta-test} domains $\breve{\mathcal{S}}$ (virtual-test domain). This is to mimic real train-test domain-shifts so that over many iterations we can train a model to achieve good \textit{generalization} in the \textbf{final-test} evaluated on target domains $\mathcal{T}$. The overall methodological flow is illustrated schematically in Fig.~\ref{illustration} and summarised in Algorithm~\ref{alg:mldg}. This model-agnostic approach can be flexibly applied to both supervised and reinforcement learning as elaborated in the following sections.

\begin{algorithm}[t]
   \caption{Meta-Learning Domain Generalization}\label{alg:mldg}
\begin{algorithmic}[1]
\Procedure{MLDG}{} 
\State \textbf{Input}: Domains $\mathcal{S}$
\State \textbf{Init}: Model parameters $\Theta$.  Hyperparameters $\alpha,\beta,\gamma$.
\For{ite \textbf{in} iterations}
\State \textbf{Split}: $\bar{\mathcal{S}}$ and $\breve{\mathcal{S}} \gets \mathcal{S}$ 
\State \textbf{Meta-train}: Gradients $\nabla_{\Theta} = \mathcal{F}_{\Theta}'( 
\bar{\mathcal{S}}; \Theta)$
\State Updated parameters $\Theta'=\Theta-\alpha \nabla_{\Theta}$ 
\State \textbf{Meta-test}: Loss is $\mathcal{G}(\breve{\mathcal{S}}; \Theta')$.
\State \textbf{Meta-optimization}: Update $\Theta$
$$
\Theta = \Theta - \gamma \frac{\partial (\mathcal{F}(\bar{\mathcal{S}}; \Theta) + \beta \mathcal{G}(\breve{\mathcal{S}}; \Theta - \alpha \nabla_\Theta))}{\partial \Theta}
$$

\EndFor
\EndProcedure
\end{algorithmic}
  \end{algorithm}

  \begin{algorithm}[t]
   \caption{MLDG for Reinforcement Learning}\label{alg:mldg-rl}
\begin{algorithmic}[1]
\Procedure{MLDG-RL}{}
\State \textbf{Input}: Environment domains $\mathcal{S}$
\State \textbf{Init}: Policy params $\Theta$, Hyperparameters $\alpha,\beta,\gamma$.
\For{ite \textbf{in} iterations}
\State \textbf{Split}: $\bar{\mathcal{S}}$ and $\breve{\mathcal{S}} \gets \mathcal{S}$ 
\State \textbf{Meta-train}: 
\State Collect trajectories $\bar{\tau}$ applying policy $\Theta$ in $\bar{\mathcal{S}}$.
\State Loss: $\mathcal{F}(\bar{\tau}, \Theta)$.
\State Gradient: $\nabla_{\Theta} = \mathcal{F}_{\Theta}'(\bar{\tau}, \Theta)$.
\State Updated parameters: $\Theta'=\Theta - \alpha \nabla_{\Theta}$.
\State \textbf{Meta-test}: 
\State Collect trajectories $\breve{\tau}$ applying policy $\Theta^{'}$ in $\breve{\mathcal{S}}$.
\State Loss $\mathcal{G}(\breve{\tau}, \Theta - \alpha\nabla_{\Theta})$.
\State \textbf{Meta-optimization}: $$\Theta = \Theta - \gamma \frac{\partial (\mathcal{F}(\bar{\tau}, \Theta) + \beta \mathcal{G}(\breve{\tau}, \Theta - \alpha \mathcal{F}'(\bar{\tau}, \Theta)))}{\partial \Theta}  $$
\EndFor
\EndProcedure
\end{algorithmic}
  \end{algorithm}
  
\subsection{Supervised Learning}

We first describe how to apply our method to supervised learning. We assume a loss function $l(\hat{y}, y)$ between the predicted and true labels $\hat{y}$ and $y$. For example in multi-class classification the cross-entropy loss: $l(\hat{y}, y) = -\hat{y} \log (y)$. The process is outlined in the steps below. 

\keypoint{Meta-Train} 
The model is updated on all the $S-V$ meta-train domains $\bar{\mathcal{S}}$ in aggregate, and the loss function is,
\begin{equation}
\label{meta-train}
\mathcal{F}(.) = \frac{1}{S-V} \sum^{S-V}_{i=1}\frac{1}{{N_i}}\sum^{N_i}_{j=1} \ell_\Theta(\hat{{y}}^{(i)}_j,{y}^{(i)}_j)
\end{equation}
\noindent where ${y}^{(i)}_j$ indicates the $j$th point among ${N_i}$ in the $i$th domain. The model is parameterized by $\Theta$, so the gradient of $\Theta$ calculated respect to this loss function is $\nabla_{\Theta}$, and optimization will update the model as $\Theta^{'} =\Theta - \alpha \nabla_{\Theta}$.

\keypoint{Meta-Test} In each mini-batch 
the model is also \textit{virtually} evaluated on the $V$ meta-test domains $\breve{\mathcal{S}}$. 
This meta-test evaluation simulates testing on new domains with different statistics, in order to allow {learning} to generalize across domains. The loss for the adapted parameters calculated on the meta-test domains is as below,
\begin{equation}
\label{meta-test}
\mathcal{G}(\cdot) = \frac{1}{V} \sum^{V}_{i=1} \frac{1}{{N}_{i}}\sum^{{N}_{i}}_{j=1} \ell_{\Theta^{'}}(\hat{{y}}_j^{(i)},{{y}}_j^{(i)})
\end{equation}
where, ${N}_{i}$ is the number samples of the $i$th meta-test domain, and the loss on the meta-test domain is calculated using the \emph{updated} parameters $\Theta^{'}$ from meta-train. \textcolor{black}{This means that for optimisation with respect to $\mathcal{G}$ we will need the second derivative with respect to $\Theta$.}


\keypoint{Summary} The meta-train and meta-test are optimised simultaneously, so the final objective is:

\begin{equation}
\underset{\Theta}{\operatorname{argmin}}~~ \mathcal{F}(\Theta) + \beta \mathcal{G}(\Theta - \alpha \mathcal{F}'(\Theta))
\label{eq:vanilla-mldg}
\end{equation}
\noindent where $\alpha$ is the meta-train step size and $\beta$ weights meta-train and meta-test. Objective (Eq.~\ref{eq:vanilla-mldg}) is itself trained by gradient descent (Alg.~\ref{alg:mldg}).

\keypoint{Final-Test} After Eq.~\ref{eq:vanilla-mldg} is optimised to convergence 
on the source domains, we deploy the final model $\Theta$ on the truly held-out target domain(s).

\subsection{Reinforcement Learning}
In application to the reinforcement learning (RL) setting, we now assume an agent with a policy $\pi$ that inputs states $x$ and produces actions $a$ in a sequential decision making task: $a_t=\pi_\Theta(x_t)$. The agent operates in an environment defined by a Markov decision process (MDP) $q(x_{t+1}|x_t,a_t)$ and its goal is to maximize its return, the (potentially discounted) sum of rewards \textcolor{black}{$\mathcal{R}=\sum_t \delta^t R_t(x_t,a_t)$}. 

While tasks in a supervised learning setting map to reward functions in an RL setting \cite{finn2017model}, domains map to solving the same task (reward function) with differences in the environment (MDP or observation function). Thus DG is to achieve an agent with improved generalization ability in the sense of maintaining ability to maximize reward when subject to changes in its operating environment (MDP) --  without being allowed any rewarded ($\approx$ supervised domain adaptation \cite{finn2017model,ammar2014pgella}), or un-rewarded ($\approx$ unsupervised domain adaptation \cite{finn2017sslRL,ammar2015unsupTransferPGRL}) trials in the target environment for adaptation. 
The key idea is still to achieve DG by simulating train-test domain shift during training. Meta-optimisation then trains for generalization across environmental conditions. The overall process is summarised in Algorithm~\ref{alg:mldg-rl} and elucidated in the steps below. Note that the MLDG strategy can be straightforwardly applied on-policy with policy-gradient (PG) {\cite{williams1992reinforce}, or off-policy with Q-learning \cite{mnih2015dqn}. For simplicity we describe the PG variant.


\keypoint{Meta-train:} 
In meta-training, the loss function $\mathcal{F}(\cdot)$ now corresponds to the negative return $\mathcal{R}$ of policy $\pi_\Theta$, averaged over all the meta-training environments in $\bar{\mathcal{S}}$. \textcolor{black}{Update of the policy parameters  $\Theta$ is performed by REINFORCE \cite{williams1992reinforce} (or Q-learning \cite{mnih2015dqn}), leading to updated parameters $\Theta^{'}$}.

\keypoint{Meta-test:} Similarly to the SL approach, we now evaluate the model on $V$ meta-test domains $\breve{S}$. The meta-test loss $\mathcal{G}$ is again the average negative return on meta-test environments. For RL calculating this loss requires rolling out the meta-train updated policy $\Theta^{'}$ in the meta-test domains to collect new trajectories and rewards. 





\subsection{Analysis of MLDG}
In this section we provide some analysis to help better understand the proposed  method. The objective of MLDG is:
\begin{equation}
\underset{\Theta}{\operatorname{argmin}}~~ \mathcal{F}(\Theta) + \beta \mathcal{G}(\Theta - \alpha \mathcal{F}'(\Theta))
\label{vanilla-mldg}
\end{equation}
where $\mathcal{F}(.)$ is the loss from the aggregated meta-train domains (Eq.~\ref{meta-train}),  $\mathcal{G}(.)$ is the loss from the aggregated meta-test domains (Eq.~\ref{meta-test}), and $\mathcal{F}'(\Theta)$ is the gradient of the training loss $\mathcal{F}(\Theta)$ w.r.t `$\Theta$'.
This can be understood as: ``\emph{tune such that after updating the meta-train domains, performance is also good on the meta-test domains}''. 

For another perspective on the MLDG objective, we can do the first order Taylor expansion for the second term, i.e. 
\begin{equation}
\mathcal{G}(x)=\mathcal{G}(\dot{x})+\mathcal{G}'(\dot{x})\times(x-\dot{x})
\end{equation}
where $\dot{x}$ is an arbitrary point that is close to $x$. The multi-variable form – $x$ is a vector and $\mathcal{G}(x)$ is a scalar.

Assume we have $x=\Theta - \alpha \mathcal{F}'(\Theta)$, and we choose the $\dot{x}$ to be $\Theta$. Then, we have 
\begin{equation}
\mathcal{G}(\Theta - \alpha \mathcal{F}'(\Theta)) = \mathcal{G}(\Theta) + \mathcal{G}'(\Theta)\cdot(-\alpha \mathcal{F}'(\Theta))
\end{equation}

\noindent and the objective function becomes 
\begin{equation}
\underset{\Theta}{\operatorname{argmin}}~~ \mathcal{F}(\Theta) + \beta \mathcal{G}(\Theta) - \beta\alpha(\mathcal{G}'(\Theta)\cdot \mathcal{F}'(\Theta)).\label{eq:alignGrad}
\end{equation}

This reveals that we want to: (i) \textit{minimize} the loss on both meta-train and meta-test domains, and (ii) \textit{maximize} the dot product of $\mathcal{G}'(\Theta)$ and $\mathcal{F}'(\Theta)$. Minimizing the loss on both domains (i) is intuitive. To understand (ii), recall the dot operation computes the similarity of two vectors: $a\cdot b = ||a||_{2}||b||_{2}\cos (\delta)$, where $\delta$ is the angle between  vectors $a$ and $b$. If $a$ and $b$ are unit normalized, this computes cosine similarity exactly. Though $\mathcal{G}'(\Theta)$ and $\mathcal{F}'(\Theta)$ are not normalized, the dot product is still larger if 
these vectors are in a {similar direction}. 

Since $\mathcal{G}'(\Theta)$ and $\mathcal{F}'(\Theta)$ are loss gradients in two sets of domains, then `similar direction' means the \emph{direction of improvement} in each set of domains is similar. Thus the overall objective can be seen as: ``\emph{tune such that both domains' losses are minimised, and also such that they descend in a coordinated way}''. \textcolor{black}{In a conventional optimisation of $\arg\min_\Theta \mathcal{F}(\Theta)+\mathcal{G}(\Theta)$, there is no such constraint on coordination. The optimiser will happily tune asymmetrically, e.g., focusing on which ever domain is easier to minimise.} \textcolor{black}{The regularisation provided by the third term in  Eq.~\ref{eq:alignGrad}  prefers updates to weights where the two optimisation surfaces agree on the gradient. It reduces overfitting to a single domain by finding a route to minimisation where both sub-problems agree on the direction at all points along the route}.

\subsection{Alternative Variants of MLDG}
Based on the discussion above, we propose some variants inspired by the vanilla MLDG method. 
Variant \textbf{MLDG-GC} in Eq.~\ref{mldg-gradnormsim} is based on the Taylor expansion and gradient alignment intution discussed earlier -- with the regularizer updated to normalize the gradients so that it indeed computes cosine similarity. 
\begin{equation}
\underset{\Theta}{\operatorname{argmin}}~~ \mathcal{F}(\Theta) + \beta \mathcal{G}(\Theta) - \beta\alpha \frac{\mathcal{F}'(\Theta) \cdot \mathcal{G}'(\Theta)}{\|\mathcal{F}'(\Theta)\|_2 \|\mathcal{G}'(\Theta)\|_2}
\label{mldg-gradnormsim}
\end{equation}
\textcolor{black}{Another perspective on `similar direction' gradients is that once meta-train has converged, you also no longer need to update the parameters on the meta-test domains. I.e., at a good solution, meta-test gradients are close to zero. With this intuition variant \textbf{MLDG-GN} is proposed in Eq.~\ref{mldg-gn}.}
\begin{equation}
\underset{\Theta}{\operatorname{argmin}}~~ \mathcal{F}(\Theta) + \beta \|\mathcal{G}'(\Theta - \alpha \mathcal{F}'(\Theta))\|^2_2
\label{mldg-gn}
\end{equation}
Clearly MLDG-GN needs a good initialisation to be reasonable, so we initialise MLDG-GN with the domain aggregation baseline. In the experiments section we will compare these alternative variants to the initially proposed MLDG.

\section{Experiments}
To evaluate our method, we compare it with various alternatives on four different problems, including an illustrative synthetic experiment, a challenging recent computer vision benchmark for multi-class classification across different domains, and two classic reinforcement learning problems, Cart-Pole and Mountain Car. In each case we compare to the baseline of aggregating the data from all source domains to train a single model that ignores domains entirely, as well as various alternative DG methods. \textcolor{black}{As shown in \cite{da2017dg}, the former simple baseline can be very effective and outperform many purpose designed DG models.}

\subsection{Experiment \rom{1}: Illustrative Synthetic Experiment}
To illustrate our approach, we construct a synthetic binary classification experiment. We synthesize nine domains by sampling curved deviations from a diagonal line classifier. We treat eight of these as sources for meta-learning and hold out the last for final-test. Fig. \ref{toy} shows the nine synthetic domains which are related in form but differ in the details of their decision boundary. A one-hidden layer MLP (50 hidden neurons, RELU activation) is used as the base classifier. 


\vspace{0.1cm}\noindent \textbf{Baselines:} \quad \textbf{MLP-All}: Simple baseline of aggregating all source domains for training. \textbf{MLDG}: Our main proposed \name~ method (Eq. \ref{vanilla-mldg}). 
\textbf{MLDG-GC} and \textbf{MLDG-GN}: variants of our method in 
Eq. \ref{mldg-gradnormsim} and Eq. \ref{mldg-gn} respectively.

\begin{figure}[t]
\centering
\begin{subfigure}{1.0\columnwidth}
\includegraphics[trim = 10mm 6mm 10mm 6mm, clip,width=.32\columnwidth]{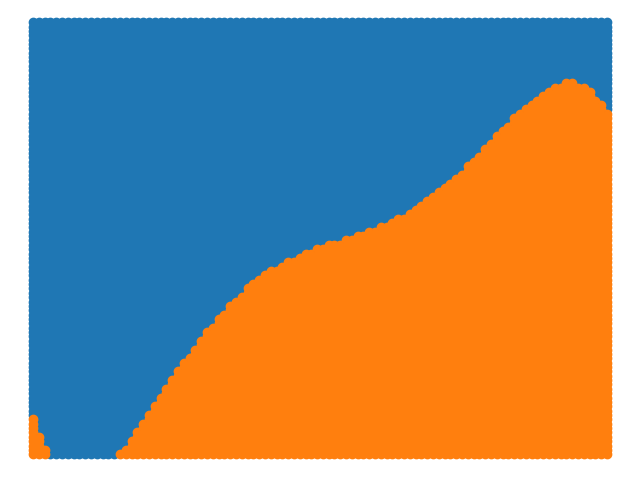}
\includegraphics[trim = 10mm 6mm 10mm 6mm, clip,width=.32\columnwidth]{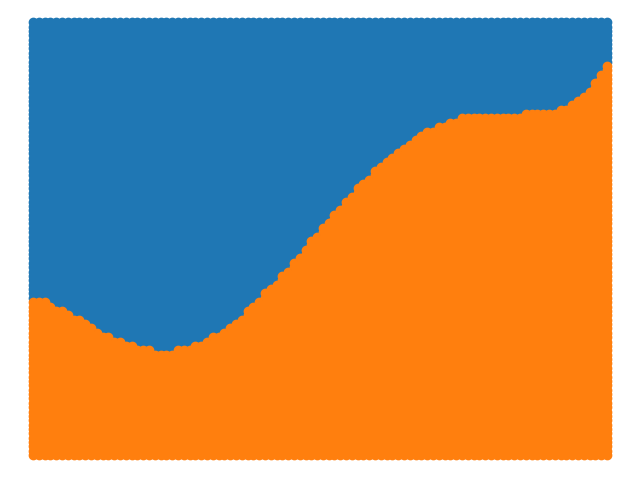}
\includegraphics[trim = 10mm 6mm 10mm 6mm, clip,width=.32\columnwidth]{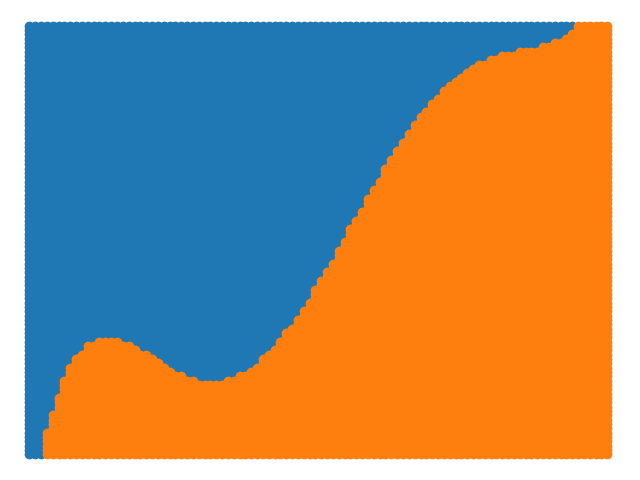}

\includegraphics[trim = 10mm 6mm 10mm 6mm, clip,width=.32\columnwidth]{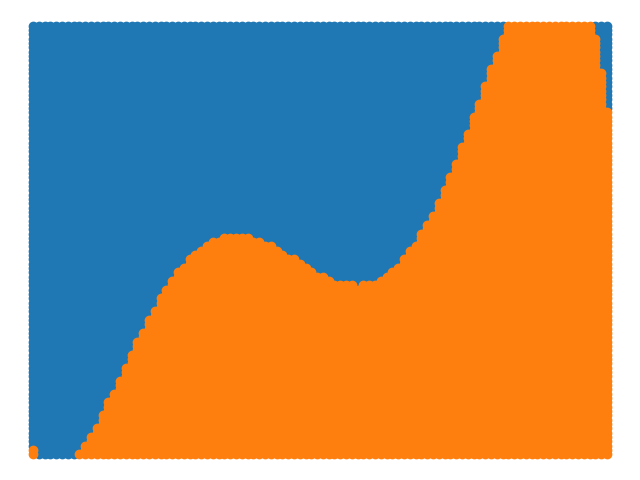}
\includegraphics[trim = 10mm 6mm 10mm 6mm, clip,width=.32\columnwidth]{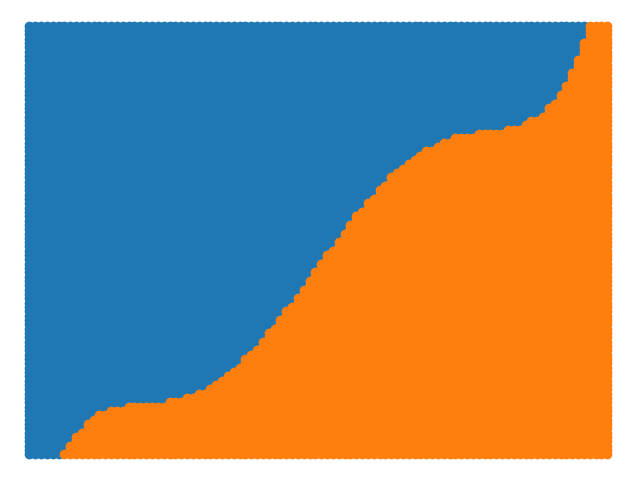}
\includegraphics[trim = 10mm 6mm 10mm 6mm, clip,width=.32\columnwidth]{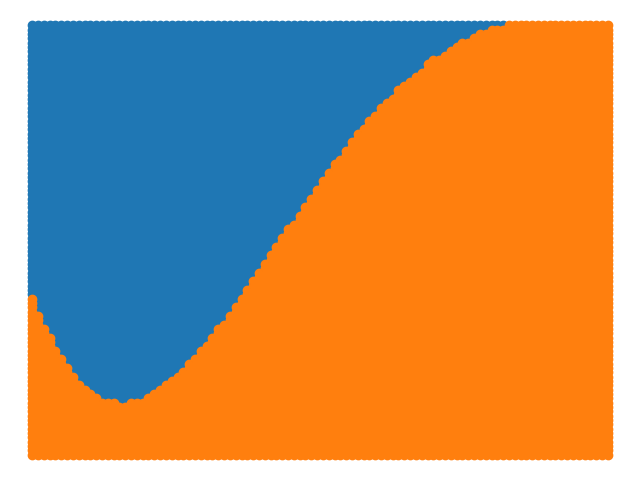}

\includegraphics[trim = 10mm 6mm 10mm 6mm, clip,width=.32\columnwidth]{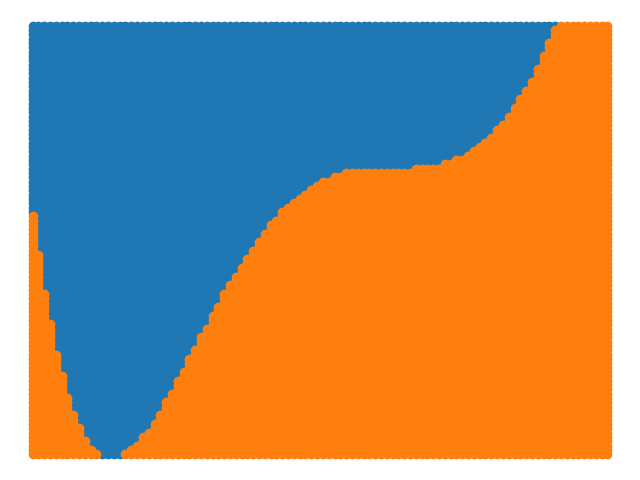}
\includegraphics[trim = 10mm 6mm 10mm 6mm, clip,width=.32\columnwidth]{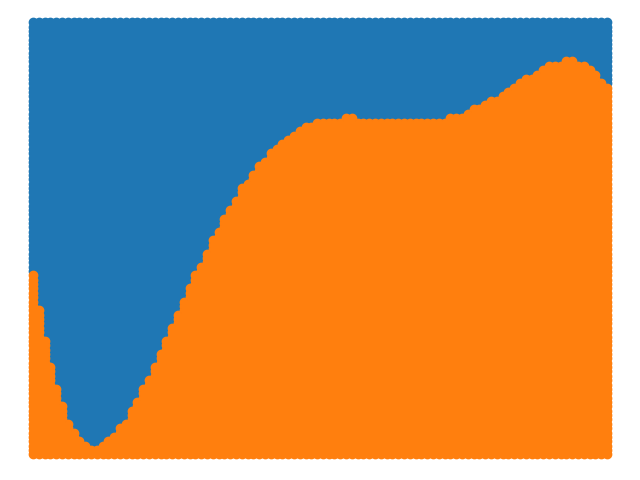}
\includegraphics[trim = 10mm 6mm 10mm 6mm, clip, width=.32\columnwidth]{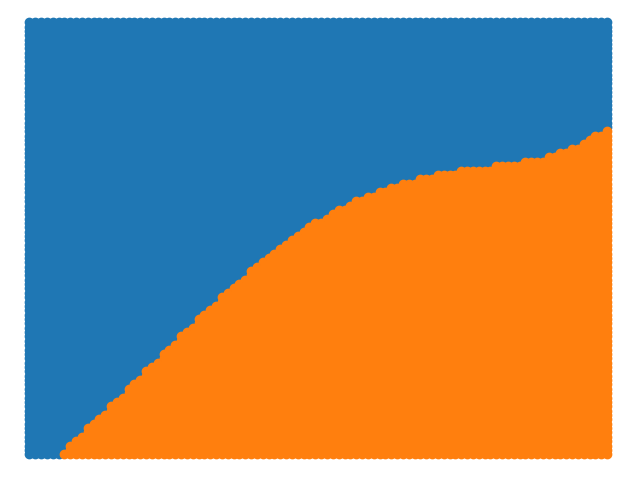}
\caption{Synthetic training domains for binary classification}
\label{toy}
\end{subfigure}
\begin{subfigure}{1.0\columnwidth}
\centering
\includegraphics[trim = 8mm 5mm 8mm 5mm, clip, width=.24\columnwidth]{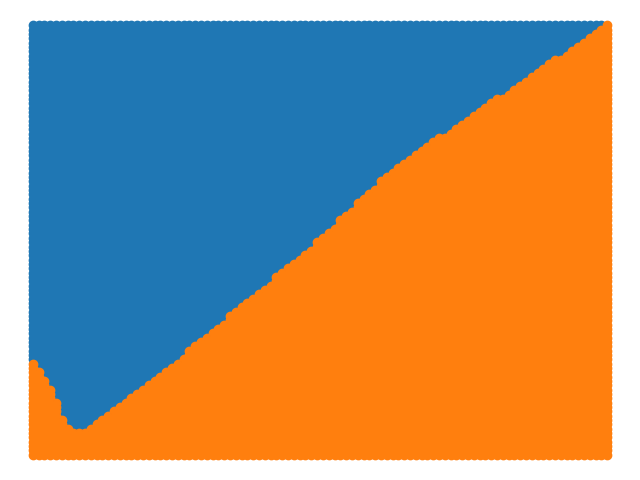}
\includegraphics[trim = 8mm 5mm 8mm 5mm, clip, width=.24\columnwidth]{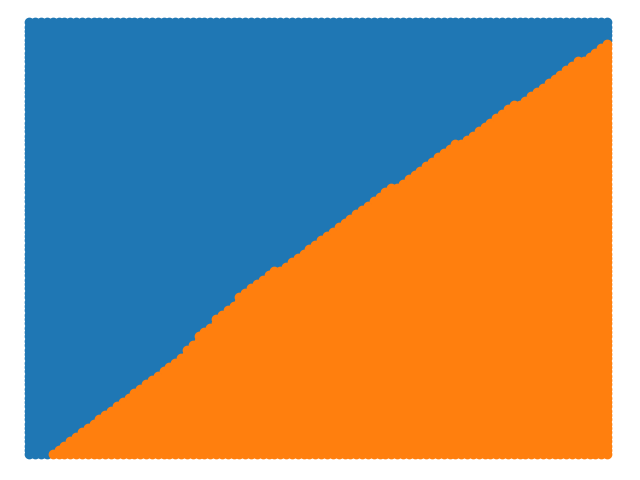}
\includegraphics[trim = 8mm 5mm 8mm 5mm, clip, width=.24\columnwidth]{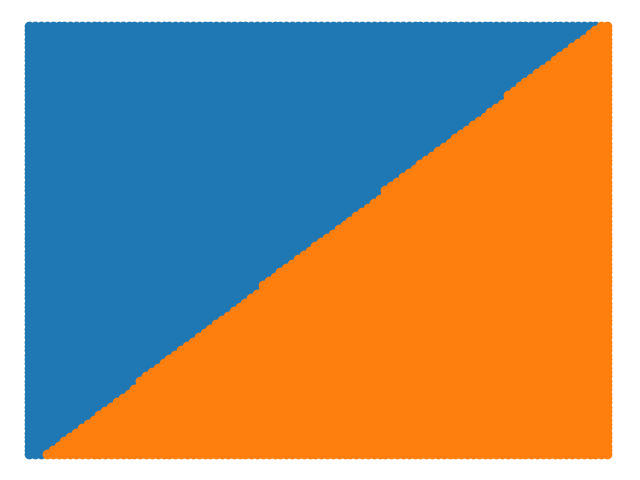}
\includegraphics[trim = 8mm 5mm 8mm 5mm, clip, width=.24\columnwidth]{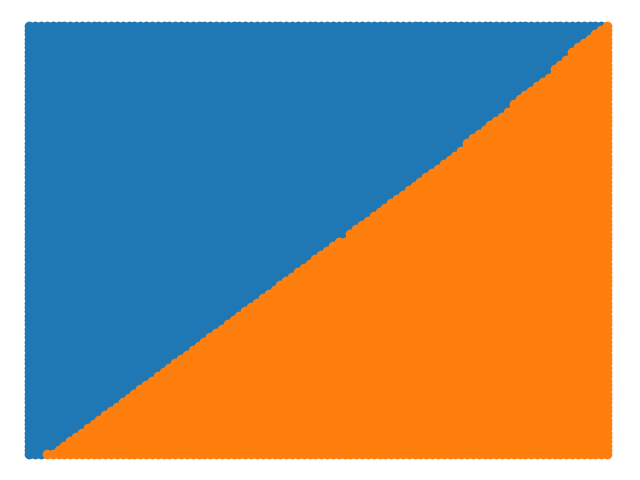}
\caption{Learned decision boundaries. From left to right: MLP-All; MLDG; MLDG-GC; MLDG-GN.}
\end{subfigure}
\caption{Synthetic experiment illustrating MLDG.}
\label{toy-res}
\end{figure}

\vspace{0.1cm}\noindent \textbf{Results:}\quad From the results Fig. \ref{toy-res} we can see that the baseline MLP-ALL over-fits on the training domains.  Despite aggregating eight sources, it fits a curve in the bottom left corner rather than the underlying diagonal line. Our methods all draw nearly straight lines. 
These results illustrate that the MLDG approach helps to avoid overfitting to specific source domains and learn a more generalizable model.


\subsection{Experiment \rom{2}: Object Recognition}
We next evaluate the efficacy of MLDG on a recent challenging object recognition DG task in computer vision. Specifically, we used the PACS multi-domain recognition benchmark, a new dataset designed for the cross-domain recognition problems \cite{da2017dg}\footnote{\url{http://sketchx.eecs.qmul.ac.uk}}. This dataset has 9991 images in total across 7 categories (`dog', `elephant', `giraffe', `guitar', `house', `horse' and `person') and 4 domains of different stylistic depictions (`Photo', `Art painting', `Cartoon' and `Sketch'). The diverse depiction styles provide a significant domain gap. The goal is to train in set of domains and recognise objects in a disjoint domain. E.g., recognise photos given only various artistic depictions for training.


\vspace{0.1cm}\noindent \textbf{Baselines:}\quad We use the ImageNet pre-trained AlexNet CNN \cite{krizhevsky2012imagenetDeepCNN} as the base network in each competitor for fair comparison, and compare the following models: \textbf{D-MTAE}: a multi-task auto encoder designed for the DG problems \cite{ghifary2015domain}. \textbf{Deep-All}: Vanilla AlexNet trained on the aggregation of data from all source domains. This baseline that outperforms many prior DG methods as presented in  \cite{da2017dg}. 
\textbf{DSN}: The domain separation network learns specific and shared representation components for the source and target domains \cite{bousmalis2016domain}. We re-purpose the original DSN from the domain adaptation to the DG task. \textbf{AlexNet+TF}: the low-rank paramaterized network provides prior state of the art on this benchmark \cite{da2017dg}.


\keypoint{Settings:} We implement MLDG in tensorflow. We use SGD optimiser with learning rate $5e-4$ (exponential decay is used with decay step $15k$ and decay rate $0.96$) and mini-batch 64. 
Meanwhile, parameters $\alpha,\beta,\gamma$ are set to $5e-4, 1.0$ and $5e-4$. For final-test, we use the best performing model on the validation set after 45$k$ iterations.

\begin{table*}[t]
\centering
\caption{Cross-domain recognition accuracy (Multi-class accuracy) on the PACS dataset. Best performance in bold.}
\label{pacs}
\scalebox{0.9}{
\begin{tabular}{cccccc}
\hline
               & D-MTAE \cite{ghifary2015domain} & Deep-all & DSN \cite{bousmalis2016domain} & AlexNet+TF \cite{da2017dg}  & MLDG (CNN)\\
              \hline
art\_painting  & 60.27 & 64.91 &61.13 &62.86 &     \textbf{66.23}                \\
cartoon        & 58.65 & 64.28 &66.54 &\textbf{66.97} & 	   66.88			  \\
photo          & \textbf{91.12} & 86.67 &83.25 &89.50 &     88.00                \\
sketch         & 47.86 & 53.08 &58.58 &57.51 &     \textbf{58.96}                \\
\hline
Ave.           & 64.48 & 67.24 &67.37 &69.21 &    \textbf{70.01}               \\
\hline
\end{tabular}
}
\end{table*}

\begin{table}[t]
\centering
\caption{PACS benchmark: Ablation study of MLDG.}
\label{eva-imp-mldg}
\resizebox{1.0\columnwidth}{!}{
\begin{tabular}{ccccc}
\hline
& Deep-All & MLDG ($\alpha=0$) & MLDG (FC) & MLDG (CNN) \\
\hline
art\_painting &64.91& 64.37 &65.54& 66.23\\
cartoon       &64.28& 65.39 &66.37& 66.88\\
photo         &86.67& 86.67 &88.30& 88.00\\
sketch        &53.08& 55.29 &55.34& 58.96\\
\hline
Ave. &67.24 & 67.93 &68.89&70.01\\
\hline
\end{tabular}
}
\end{table}

\begin{table}[t]
\centering
\caption{PACS benchmark: Evaluation of MLDG variants.}
\label{eva-variants-mldg}
\resizebox{1.0\columnwidth}{!}{
\begin{tabular}{cccc}
\hline
& Deep-All & MLDG-GC (Eq.~\ref{mldg-gradnormsim}) & MLDG-GN (Eq.~\ref{mldg-gn}) \\
\hline
art\_painting &64.91& 64.71 & 63.64 \\
cartoon       &64.28& 65.30 & 63.47  \\
photo         &86.67& 86.79 & 87.88 \\
sketch        &53.08& 56.92 & 54.94 \\
\hline
Ave. &67.24& 68.43& 67.48\\
\hline
\end{tabular}
}
\end{table}

\keypoint{Results:}  The comparison with state of the art on the PACS benchmark is shown in Table~\ref{pacs}. From the results, we can see that  MLDG surpasses the other baselines including the best prior method AlexNet+TF \cite{da2017dg}. We note that this good performance is achieved without any special architecture design  and without growing the size of the model in proportion to the number of domains (both of which are required in each of D-MTAE, DSN, and AlexNet+TF). This illustrates the flexibility of MLDG, and also highlights that its scalability compared to alternatives. AlexNet+TF for example requires approximately 2GB of memory per domain with batch size 64, meaning that it cannot be applied to more than 5 source domains on a contemporary GPU.

\keypoint{Analysis of MLDG learning:}  We next perform some ablation experiments to understand: (i) whether it is important to use MLDG end-to-end way within a CNN, and (ii) verify the impact of the meta-optimisation strategy specifically. 

To answer the first question of \emph{where} it is important to employ MLDG learning, we compare the variant \textbf{MLDG (FC)}: Only apply MLDG learning on the {FC} layers of AlexNet. This is in contrast to our full model \textbf{MLDG (CNN) }, which applies learning to all layers of AlexNet. Comparing MLDG (FC) to vanilla Deep-All AlexNet in Table~\ref{eva-imp-mldg}, we see a benefit of $\approx 1.6\%$ is obtained by  MLDG learning on the FC layers. Comparing full MLDG we see that a further $\approx 1.1\%$ benefit is obtained by applying MLDG learning to the convolutional layers, for a total of $\approx 2.7\%$ margin over Deep-All.

To verify the impact of the meta-optimisation strategy, we apply MLDG with setting $\alpha=0$, in which case the objective is merely the sum of the training and validation (meta-test) domains' losses. From the results in  Table~\ref{eva-imp-mldg}, we see that it performs comparably with Deep-All. Thus the key benefit of MLDG is indeed in the meta-optimisation step. 

\keypoint{Analysis of MLDG variants:} In the Table \ref{eva-variants-mldg}, the original MLDG method is compared to the two variants also proposed in the methodology. 
In this experiment we found that while the MLDG-GC (cosine) and MLDG-GN (gradient norm) variants provide some benefit compared to Deep-All, the vanilla MLDG performs best.



\subsection{Experiment \rom{3}: Cart-Pole}
We next demonstrate that MLDG also applies to RL problems. First we study the classic Cart Pole problem \cite{openaigym}. The objective is to balance a pole upright by moving a cart. The action space is discrete --  left or right. The state it has four elements: the position and velocity of cart and angular position and velocity of the pole. 

\keypoint{Settings:} We perform two sub-experiments by modifying the OpenAI Gym simulator to provide environments with different properties. In the first we vary one domain factor by changing the pole length. We simulate 9 domains with pole lengths $[0.5,1.0,\dots,4.5]$.  In the second we vary multiple domain factors -- pole length $[0.5, 2.5, 4.5]$ and cart mass $[1, 2, 3]$. In both experiments we randomly choose 6 source domains for training and hold out 3 domains for (true) testing. Since the game can last forever if the pole does not fall, we cap the maximum steps to $200$. \textcolor{black}{We train on the observed domains for 500 games per domain.}
Then, for each held-out domain, we play $500$ games, and report the average reward. For fair comparison, the policy architecture for all models is a 1-hidden layer neural network with 50 hidden units. The reward structure is +1 for each time-step the pole is successfully balanced, so the maximum reward is $200$. All methods are trained with vanilla REINFORCE policy gradient \cite{williams1992reinforce}. 

\keypoint{Baselines:} We compare the following alternative approaches:  \textbf{RL-All}: The reinforcement-learning analogy to `Deep-ALL' in the recognition experiment. Trains a single policy by aggregating the reward from all six source domains. \textbf{RL-Random-Source}: Different from RL-All, it trains on a single randomly selected source domain. Total training trials are controlled so it gets the same number of trials in one domain as RL-All gets in multiple domains.
\textbf{RL-Undobias}: Adaptation of the (linear) undo-bias model of \cite{khosla2012undoing} updated to non-linear multi-layer network as per \cite{da2017dg}. The neural network is trained to factor domain-specific and a single domain-agnostic components on six source domains. The domain agnostic component is then transferred for testing on held out final-testing domains. \textbf{RL-MLDG}: Our  MLDG. \textbf{RL-MLDG-GC}: Our MLDG variant. \textbf{RL-MLDG-GN}: Our MLDG variant.  In each mini-batch, we split the $S=6$ source domains into $V=2$ meta-test  and $S-V=4$ meta-train domains.

\begin{table}[t]
\centering
\caption{Cart-Pole RL. Domain generalisation performance across pole length. Average reward testing on 3 held out domains with random lengths. Upper bound: 200.}
\label{rl}
\resizebox{1.0\columnwidth}{!}{
\begin{tabular}{cccc}
\hline
      Method      & RL-Random-Source  &  RL-All & RL-Undobias  \\
      Return       & $133.74 \pm 6.79$ & $97.39\pm 73.49$ & $113.52\pm 11.65$ \\
      \hline
  Method      & RL-MLDG & RL-MLDG-GC & RL-MLDG-GN \\
   Return   & $165.34\pm 3.38$ & $129.56\pm 2.51$ &  $175.25 \pm 3.16$ \\
              \hline
\end{tabular}
}
\end{table}

\begin{table}[t]
\centering
\caption{Cart-Pole RL. Generalisation performance across both pole length and cart mass. Return testing on 3 held out domains with random length and mass. Upper bound: 200.}
\label{cartpole-polelen-cartmass}
\resizebox{1.0\columnwidth}{!}{
\begin{tabular}{cccc}
\hline
      Method      & RL-Random-Source  &  RL-All & RL-Undobias  \\
      Return       & $98.22 \pm 20.35$ & $144.21 \pm 9.23$  & $150.46\pm 17.59$  \\
      \hline
  Method      & RL-MLDG & RL-MLDG-GC & RL-MLDG-GN \\
   Return   &  $170.81 \pm 9.90$& $147.76\pm 4.41$  & $164.97\pm 8.45$  \\
              \hline
\end{tabular}
}
\end{table}

\keypoint{Results:} All experiments are repeated 10 times to reduce the impact of specific observed/held-out domain sampling. From the results in Tables~\ref{rl} and \ref{cartpole-polelen-cartmass}, \textcolor{black}{we  see the impact of domain shift. No methods reach $200$ (upper bound given the length cap) for  unseen domains reliably. However, the proposed MLDG provides the best domain generalization} and  significantly outperform the baselines. It is interesting to note that RL-Random-Source outperforms RL-All \textcolor{black}{in Table~\ref{rl}}, which is different than in vision problems where simply aggregating more domains is usually a reasonable strategy. Although RL-All is exposed to more diverse data, learning a single policy by naively `averaging' over rewards for multiple distinct problems can sometimes be detrimental \cite{yang2017metacritic}, 

\keypoint{Analysis of MLDG variants:} \textcolor{black}{Comparing MLDG with its variants MLDG-GC and MLDG-GN we found that MLDG-GN is comparable to vanilla MLDG on this problem, while MLDG-GC is slightly worse. }

\subsection{Experiment \rom{4}: Mountain Car}

Our second RL experiment is the classic mountain car problem \cite{openaigym}. The car is positioned between two mountains, and the agent needs to drive the car (back or forth) so that it can hit the peak of the right mountain. The difficulty of this problem is that the car engine is not strong enough to drive up the right mountain directly. The agent has to figure out a solution of driving up the left mountain to first generate momentum before driving up the right mountain. The state observation in this game consists two elements: the position and velocity of the car. There are three available actions: drive left, do nothing, and drive right.

\keypoint{Settings:} We simulate domain bias by randomly drawing the height of the mountains in each domain. Similar to cartpole, we simulate 9 domains in total, and 3 domains are held-out. 
In contrast to cartpole, it is very difficult for a random policy to finish a full game, as it is likely to be stuck forever. Thus instead of policy gradient, we use Q learning \cite{qlearningwatkins1992} for this problem as the base RL algorithm, more specially DQNs \cite{mnih2015dqn}. For held out domains we play 100 games each {without updating}. 
\textcolor{black}{The reward structure is -1 each time step before reaching the target.} \textcolor{black}{The Q-network is again a 1 hidden layer MLP.} 

\keypoint{Baselines:} We evaluate the following alternatives \quad \textbf{RL-Random-Source}: Trains a single policy on one random source domain. \textbf{RL-All}: Trains a single policy on 6 source domains in aggregation. \textbf{RL-Undobias}: DG parametrised Q-network adaptation of \cite{khosla2012undoing,da2017dg} as per cart-pole. \textbf{RL-MLDG}: Our  MLDG. And its variants \textbf{RL-MLDG-GC} and \textbf{RL-MLDG-GN}. In each mini-batch, we split the $S=6$ source domains into $V=2$ meta-test domains, and $S-V=4$ meta-train domains.

\begin{table}[t]
\centering
\caption{Domain generalisation performance for mountain car. Failure rate ($\downarrow$) and reward ($\uparrow$) on held out testing domains with random mountain heights.}
\label{rl2}
\resizebox{1.0\columnwidth}{!}{
\begin{tabular}{cccc}
\hline
      Mountain Car      & RL-Random-Source &  RL-All & RL-Undobias  \\
      \hline
      Avg. F Rate       & $0.55\pm 0.07$ &$0.05\pm 0.02$ & $0.08\pm 0.04$ \\
      Avg. Return       & $-191.07\pm 3.01$&$-141.35\pm 2.64$ & $-124.48\pm 3.22$ \\
              \hline
       Mountain Car       & RL-MLDG & RL-MLDG-GC & RL-MLDG-GN \\
\hline
       Avg. F Rate        & $0.05\pm 0.02$ & $0.0 \pm 0.0$ & $1.0\pm 0.0$ \\
       Avg. Return		 & $-125.73\pm 2.76$ & $ -311.80 \pm 3.92$ & - \\
       \hline
\end{tabular}
}
\end{table}

\keypoint{Results:} All experiments are repeated 10 times to reduce the impact of random observed/held-out domain splits. From the results in Table \ref{rl2}, we again observe the performance drops from observed domains and held-out domains.
\textcolor{black}{In this benchmark, succeeding within 110 steps is a good outcome. So a reward of -110 is a good score for within domain evaluation. I.e., in the absence of domain shift.} Since it is possible for an agent to never succeed on this benchmark, particularly when testing in a distinct domain from training, we apply a limit of $20,000$ steps maximum. 
For DG testing, most methods have some failed trials ($>20,000$ steps) in final-test. 
The average reward is calculated by ignoring those failed cases. Therefore we report both failure rate and the average reward (negative time to success) in the successful cases. 
\textcolor{black}{The results show that our vanilla MLDG method outperforms the alternatives: (i) Its average reward is better than RL-All and similar to RL-UndoBias. However (ii) its fail rate is lower than RL-UndoBias. Unlike Cartpole here RL-All is more effective than Random-Source.}

\keypoint{Analysis of MLDG variants:} Only vanilla MLDG performed well here. MLDG-GC had low failure rate but low return, while MLDG-GN had very high failure rate. 

\subsection{Discussion}
The experiments show that MLDG-based meta-learning can effectively alleviate domain-shift in diverse problems including supervised and re-reinforcement learning scenarios. Whether training on the aggregate of multiple source domains was a good strategy turned out to be problem dependent (yes for PACS vision benchmark and mountain car, but not for cart pole). The extended variants of the MLDG model MLDG-GC (explicit gradient direction alignment) and MLDG-GN (gradient norm) also had mixed results with MLDG-GC performing second best on PACS, but MLDG-GN performing best on cartpole. Nevertheless the core MLDG strategy was highly effective across all problems and always outperformed prior alternatives. 

We note that studies have used the terms `domain' and `task' in different ways \cite{ammar2015crossDomainTransfer,csurka2017domainAdaptationBook}. Some problems we solved here (e.g., poles of different length) have been termed `tasks' in other studies \cite{ammar2014pgella,ammar2015crossDomainTransfer}, which would use `domain' to refer to cartpole versus mountaincar. We use the term domain in the sense of the pattern recognition community \cite{csurka2017domainAdaptationBook}, where one can learn a model with better `cross domain generalisation'. E.g. a recognition model that is robust to recognising photos vs sketches; or a policy that is more robust being deployed with poles of a different length than it was trained on. Note that if parameters like pole-length were \emph{observed}, this would be a `paramaterised' or `contextual' policy situation - for which methods already exist \cite{kupcsik2013optionGeneralization}. But in our case what meta-learning has achieved is to learn a policy that is robust to (i.e., obtains high reward despite of) hidden changes in the underlying MDP. For example balancing poles of diverse but unknown lengths.

\section{Conclusion}

We proposed a meta-learning algorithm for domain generalisation. Our method trains for domain generalisation by meta-optimisation on simulated train/test splits with domain-shift. Unlike prior model-based domain generalisation approaches, it scales well with number of domains. It is model agnostic so can be applied to different base network types, and both to supervised and reinforcement learning problems. Experimental evaluation shows state of the art results on a recent challenging visual recognition benchmark and promising results on multiple classic RL problems. 

\clearpage
{

}

\clearpage

\end{document}